\documentclass[letterpaper]{article}
\usepackage[preprint]{aaai2027}

\usepackage[hyphens]{url}
\usepackage{graphicx}
\usepackage{natbib}
\usepackage{caption}
\usepackage{algorithm}
\usepackage{algorithmic}

\usepackage{amsthm}
\usepackage{amsmath}
\usepackage{amssymb}
\usepackage{booktabs}
\usepackage{multirow}
\usepackage{mathtools}
\usepackage{nicefrac}
\usepackage{bm}
\usepackage[table]{xcolor}

\newcommand{\R}{\mathbb{R}}
\newcommand{\E}{\mathbb{E}}
\newcommand{\Cov}{\operatorname{Cov}}

\usepackage{newfloat}
\usepackage{listings}
\DeclareCaptionStyle{ruled}{labelfont=normalfont,labelsep=colon,strut=off}
\floatstyle{ruled}
\newfloat{listing}{tb}{lst}{}
\floatname{listing}{Listing}

\title{Uncertainty in a Single Pass: A Closed-Form Identity \\ for One-Step Flow Matching}

\author{
    Li Lv\textsuperscript{\rm 1},
    Jiarui Xing\textsuperscript{\rm 2},
    Xinyan Liang\textsuperscript{\rm 1},
    Jian Wang\textsuperscript{\rm 3}
}
\affiliations{
    \textsuperscript{\rm 1}Shanxi University\\
    \textsuperscript{\rm 2}Yale University\\
    \textsuperscript{\rm 3}Harvard Medical School\\
    lvlvlvli0778@gmail.com, jiarui.xing@yale.edu,\\
    liangxinyan48@163.com, jianbljh@gmail.com
}

\begin{document}

\maketitle

\begin{abstract}
Flow matching provides a highly effective framework for generative modeling, yet estimating the uncertainty of its generated samples remains a fundamental challenge. Existing methods rely on auxiliary variance heads, model ensembles, or iterative covariance propagation, which invariably introduce accumulated errors and structural interference along the noise-to-image trajectory. We resolve these limitations by adapting Tweedie's formula to the linear flow matching interpolant, yielding an exact, closed-form identity for the posterior covariance. Because this identity relies exclusively on the divergence of the learned velocity field, it can be evaluated post hoc on pretrained models without any architectural modifications. This analytical formulation proves exceptionally powerful for single-step generative frameworks. By computing the end-to-end posterior covariance in one forward pass, our approach completely bypasses the compounding errors inherent to sequential integration. This mechanism fundamentally enhances robustness, providing highly accurate and stable uncertainty estimates while preventing the numerical degradation typical of multi-step approximations. We validate our framework across benchmark datasets, including CIFAR-10 and real brain MRI scans. The resulting divergence-based uncertainty maps are highly interpretable and tightly correlated with empirical reconstruction errors. Crucially, in medical imaging, this precise localization of structural ambiguity provides essential decision support for high-stakes clinical tasks such as tumor boundary delineation and neurosurgical planning.
\end{abstract}
\section{Introduction}
 
\begin{figure*}[t]
    \centering
    \includegraphics[width=0.98\linewidth]{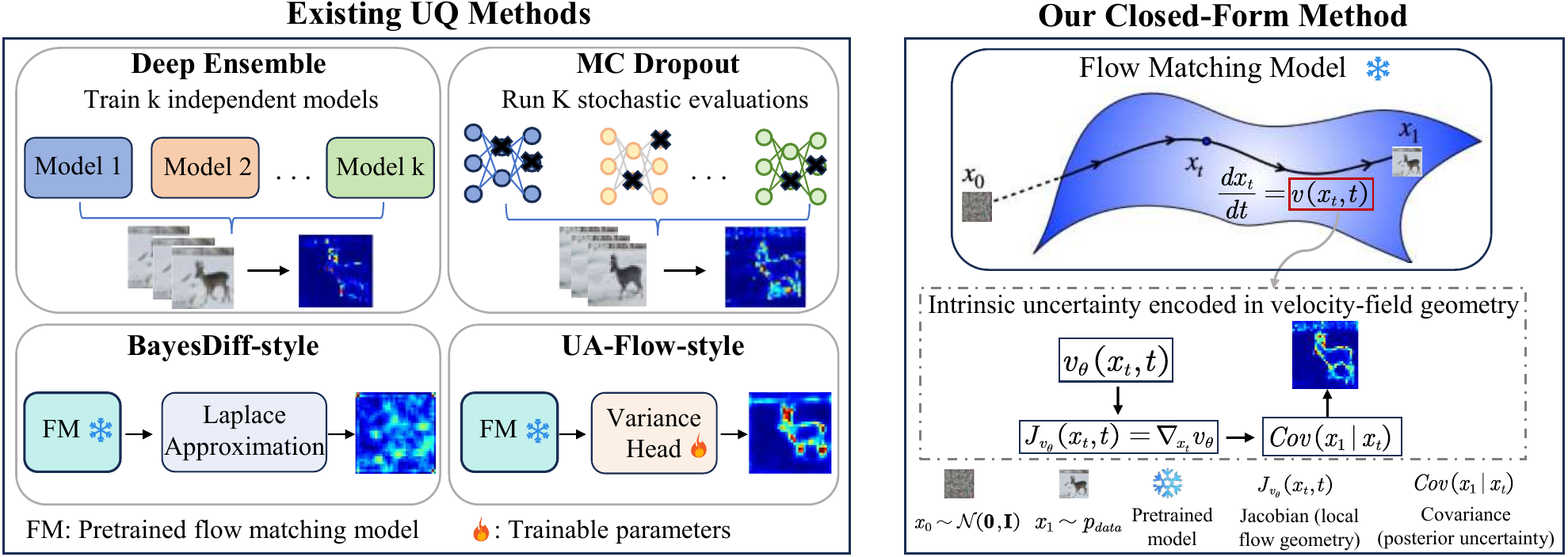}
    \caption{\textbf{Left:} Taxonomy and computational pipelines of existing uncertainty quantification methods for flow matching models. Sampling-based methods, including Deep Ensemble and MC Dropout, estimate uncertainty from prediction disagreement across independently trained models or stochastic forward passes. Auxiliary-modeling methods, including BayesDiff-style and UA-Flow-style baselines, require fitting a Bayesian last-layer posterior or training an additional variance head on a pretrained flow model. \textbf{Right:} Our closed-form uncertainty quantification method, which derives pixel-wise uncertainty directly from the local Jacobian of the pretrained velocity field, without repeated stochastic sampling, auxiliary model fitting, or additional training.}
    \label{fig:intro}
\end{figure*}
 
Flow matching~\cite{lipman2022flow,liu2023flow,albergo2023stochastic} learns a velocity field that transports samples from a simple source distribution to the data distribution. Its simulation-free training, fast inference, and nearly straight transport paths have made it a strong alternative to diffusion models in image~\cite{esser2024scaling,lin2024common,karras2024analyzing}, video~\cite{polyak2024movie}, audio, and scientific generation~\cite{liu2024imageflownet,chen2025delta}. As these models enter high-stakes applications, the reliability of individual samples becomes as important as their quality \cite{zou2023review}: a synthesized disease progression is only actionable for a clinician if it carries a confidence estimate \cite{ghesu2021quantifying}, and a generated molecular structure is easier to validate experimentally when its uncertain regions are identified in advance \cite{soleimany2021evidential}. Uncertainty quantification (UQ) \cite{he2026survey} is therefore a prerequisite for deploying flow matching in such domains, yet it remains largely unresolved.
 
Existing approaches estimate uncertainty at considerable cost. Deep ensembles~\cite{lakshminarayanan2017simple, gou2026uncertainty} require training and storing multiple independent models, which is prohibitive at the scale of modern flow matching systems. Monte Carlo dropout~\cite{gal2016dropout} requires dozens of stochastic forward passes and offers little theoretical grounding for flow-based models. UA-Flow~\cite{han2026uaflow} augments the velocity network with a heteroscedastic variance head, which requires retraining the model from scratch, and propagates the variance through the ordinary differential equation (ODE) dynamics by a first-order Taylor expansion whose error accumulates over integration steps. In the diffusion literature, BayesDiff and related methods~\cite{kou2024bayesdiff,jazbec2025generative} rely on Tweedie-style recursions that must likewise be unrolled step by step. Despite their differences, these methods share one design decision, illustrated in Fig. \ref{fig:intro}: they estimate uncertainty from the outside, treating the generative trajectory as a black box to be probed by sampling, retraining, or approximate propagation.
 
This observation raises a natural question: does the trajectory itself already carry the information needed for UQ? Under the standard linear interpolant, an intermediate state is a weighted combination of Gaussian noise and a data sample, and is therefore consistent with an entire distribution of possible endpoints. The covariance of this posterior distribution quantifies how uncertain the endpoint is given the current state, so the question becomes whether this covariance can be computed from the model itself. We show that it can. By adapting Tweedie's formula from Gaussian denoising to the flow matching interpolant, we derive a closed-form expression for the posterior covariance: up to a known time-dependent scaling, it equals the Jacobian of the learned velocity field. Closed-form posterior covariances of this kind were previously derived for diffusion models~\cite{manor2024posterior}, whose forward process adds Gaussian noise to the data; the flow matching interpolant scales signal and noise differently, and we establish the corresponding identity in the velocity-field parameterization native to flow matching. In concurrent work, FDS~\cite{cha2026training} uses the divergence of the velocity field for training-free sample refinement; although its goal is sample quality rather than uncertainty, it points to the same quantity that emerges from our analysis, suggesting that the velocity divergence is a fundamental object in the geometry of flow matching.
 
The identity translates directly into a practical method. The diagonal of the covariance yields a per-pixel uncertainty map, and its trace, which reduces to the divergence of the velocity field, yields a scalar score for the whole sample. Both are estimated with a small number of Jacobian-vector products, at a cost comparable to a few extra forward passes, and the procedure applies post hoc to pretrained models without retraining, auxiliary variance heads, or ensembles. For multi-step generation, the identity gives the endpoint uncertainty at every intermediate time; obtaining the uncertainty of the final sample would still require propagating covariance through the remaining ODE steps. One-step generators such as MeanFlow~\cite{geng2025meanflow} remove this last obstacle: because they map noise directly to the generated endpoint, applying the identity to the one-step map yields an end-to-end uncertainty estimate in a single forward pass, with no propagation at all.
 
Our contributions are threefold:

\begin{itemize}
    \item \textbf{Closed-Form UQ Formulation:} We derive an exact, parameterization-free identity for the posterior covariance under flow matching interpolants. By exposing how second-order uncertainty is intrinsically encoded within the velocity field's local geometry, our formulation enables post hoc evaluation without architectural change.
    
    \item \textbf{Single-Pass Robustness:} To our knowledge, we present the first closed-form UQ framework for single-step flow matching. Evaluating the single-pass map yields exact end-to-end uncertainty in a single forward pass, bypassing iterative propagation to eliminate error accumulation and ensure high numerical stability.
    
    \item \textbf{Empirical and Clinical Utility:} We validate our approach on CIFAR-10 and real brain MRI scans. The resulting pixel-wise uncertainty maps offer remarkable interpretability and tightly align with reconstruction errors, providing actionable spatial confidence for critical clinical workflows, including tumor boundary delineation and neurosurgical planning.
\end{itemize}
 
\section{Background}
 
\subsection{Conditional Flow Matching}
Conditional flow matching~\cite{lipman2022flow,liu2023flow} learns a velocity field that transports a source distribution $p_0 = \mathcal{N}(\mathbf{0}, \mathbf{I})$ to a target data distribution $p_1$ over $\mathbb{R}^d$, where $d$ is the data dimensionality. Given independent samples $x_0 \sim p_0$ and $x_1 \sim p_1$, one constructs the linear interpolant
\begin{equation}
x_t = t\,x_1 + (1{-}t)\,x_0, \qquad t \in [0, 1],
\label{eq:interpolant}
\end{equation}
and trains a neural network $v_\theta(x_t, t)$ to predict the conditional velocity $x_1 - x_0$ by minimizing
\begin{equation}
\mathcal{L}_{\mathrm{FM}} = \mathbb{E}_{t \sim \mathcal{U}(0,1),\, x_0,\, x_1}\bigl[\|v_\theta(x_t, t) - (x_1 - x_0)\|^2\bigr].
\label{eq:fm_loss}
\end{equation}
The property of the interpolant Eq.~\eqref{eq:interpolant} that we use throughout the paper is its conditional distribution. Since $x_0 \sim \mathcal{N}(\mathbf{0}, \mathbf{I})$ and $x_1$ is fixed, the state $x_t$ given $x_1$ is Gaussian:
\begin{equation}
p(x_t \mid x_1) = \mathcal{N}\bigl(x_t;\; t\,x_1,\; (1{-}t)^2\,\mathbf{I}\bigr).
\label{eq:cond_xt}
\end{equation}
Conditionally on the data, the intermediate state therefore behaves like a scaled data point corrupted by Gaussian noise, with signal scale $t$ and noise scale $1-t$. This Gaussianity is what makes a Tweedie-style analysis possible.
 
\subsection{Tweedie's Formula and Posterior Moments}
Tweedie's formula~\cite{robbins1956empirical,efron2011tweedie, manor2024posterior} connects the score of a noisy observation to the posterior mean of the clean signal. In its classical form, for an observation $y = x + \sigma\epsilon$ with noise level $\sigma > 0$ and $\epsilon \sim \mathcal{N}(\mathbf{0}, \mathbf{I})$, the posterior mean satisfies
\begin{equation}
\mathbb{E}[x \mid y] = y + \sigma^2\,\nabla_y \log p_\sigma(y),
\label{eq:tweedie_classic}
\end{equation}
where $p_\sigma$ denotes the marginal density of $y$. This identity underlies score-based diffusion models~\cite{song2021score,ho2020denoising}. \citet{manor2024posterior} extended it to second moments, showing that
\begin{equation}
\operatorname{Cov}(x \mid y) = \sigma^2\bigl(\sigma^2\,\nabla_y^2 \log p_\sigma(y) + \mathbf{I}\bigr),
\label{eq:tweedie_cov_classic}
\end{equation}
which links the posterior covariance to the Hessian of the log-marginal density and yields uncertainty estimates from the Jacobian of a pretrained denoiser.
 
Both identities assume the additive-noise model $y = x + \sigma\epsilon$. The flow matching interpolant Eq.~\eqref{eq:interpolant} instead scales the signal by $t$ and the noise by $1-t$, so the classical formulas do not apply directly. Deriving the analogous identities for this interpolant, in the velocity-field parameterization native to flow matching, is the subject of Section~\ref{sec:method}.
 
\subsection{MeanFlow: One-Step Generation}
MeanFlow~\cite{geng2025meanflow} replaces the instantaneous velocity in standard flow matching with a mean velocity, the average velocity over a time interval, and derives an identity that allows this quantity to be regressed directly, without numerical integration. Generation then requires a single function evaluation:
\begin{equation}
\hat{x}_1 = x_0 + \bar{u}_\theta(x_0, 0, 1),
\label{eq:meanflow_gen}
\end{equation}
where $\bar{u}_\theta$ is the learned mean-velocity network and $\hat{x}_1$ the generated sample. This yields competitive quality with a 10--100$\times$ speedup over multi-step flow matching.
 
At the population level, the posterior mean of the endpoint is related to the velocity field by
\begin{equation}
\mathbb{E}[x_1 \mid x_t] = x_t + (1{-}t)\,v_\theta(x_t, t),
\label{eq:posterior_mean}
\end{equation}
which holds for both standard flow matching, where $v_\theta$ is trained on instantaneous targets, and MeanFlow, where $\bar{u}_\theta$ is trained on mean-velocity targets, provided the network has converged. The difference is operational: standard flow matching evaluates $v_\theta$ across many steps, whereas MeanFlow evaluates $\bar{u}_\theta$ once. Our covariance identity applies to both, and the one-step setting is where its implications are most direct, as we show in Section~\ref{sec:meanflow_uq}.
\section{Method: Closed-Form Uncertainty}
\label{sec:method}

We derive the posterior covariance $\operatorname{Cov}(x_1 \mid x_t) \in \mathbb{R}^{d \times d}$ under the flow matching interpolant, where $x_1$ is the data endpoint, $x_t$ the state at time $t$, and $d$ the data dimension. Section~\ref{sec:step1} expresses the posterior mean through the marginal score; Section~\ref{sec:step2} differentiates this relation to obtain the covariance; Section~\ref{sec:step3} substitutes the velocity field, reducing the covariance to its Jacobian. Section~\ref{sec:computation} gives the estimator, and Section~\ref{sec:meanflow_uq} applies the result to one-step generators.

\subsection{Posterior Mean Identity}
\label{sec:step1}

For a fixed endpoint $x_1$, the interpolant $x_t = t x_1 + (1{-}t) x_0$ with $x_0 \sim \mathcal{N}(\mathbf{0}, \mathbf{I})$ makes $x_t$ Gaussian with mean $t x_1$ and covariance $(1{-}t)^2 \mathbf{I}$. Its log-density is
\begin{equation}
\log p(x_t \mid x_1) = -\frac{\|x_t - t x_1\|^2}{2(1{-}t)^2} + \mathrm{const},
\label{eq:cond_loglik}
\end{equation}
with conditional score
\begin{equation}
\nabla_{x_t}\log p(x_t \mid x_1) = -\frac{x_t - t x_1}{(1{-}t)^2}.
\label{eq:cond_score}
\end{equation}
The marginal score is the posterior mean of the conditional score~\cite{vincent2011connection}:
\begin{equation}
\nabla_{x_t}\log p_t(x_t) = \mathbb{E}_{x_1 \sim p(x_1 \mid x_t)}\!\bigl[\nabla_{x_t}\log p(x_t \mid x_1)\bigr].
\label{eq:score_identity}
\end{equation}
Substituting Eq.~\eqref{eq:cond_score} into Eq.~\eqref{eq:score_identity} and solving for $\mathbb{E}[x_1 \mid x_t]$ gives the Tweedie identity for the interpolant:
\begin{equation}
\mathbb{E}[x_1 \mid x_t] = \frac{1}{t}\,x_t + \frac{(1{-}t)^2}{t}\,\nabla_{x_t}\log p_t(x_t).
\label{eq:tweedie_fm}
\end{equation}
The coefficients $\nicefrac{1}{t}$ and $\nicefrac{(1{-}t)^2}{t}$ replace the single factor $\sigma^2$ of the additive-noise case Eq.~\eqref{eq:tweedie_classic}, because the interpolant scales the signal by $t$ and the noise by $1-t$.

\subsection{Posterior Covariance Identity}
\label{sec:step2}

Differentiating Eq.~\eqref{eq:tweedie_fm} with respect to $x_t$ introduces the second moment:
\begin{equation}
\nabla_{x_t}\mathbb{E}[x_1 \mid x_t] = \frac{1}{t}\,\mathbf{I} + \frac{(1{-}t)^2}{t}\,\nabla_{x_t}^2 \log p_t(x_t),
\label{eq:jacobian_pmean}
\end{equation}
where $\nabla_{x_t}^2 \log p_t(x_t)$ is the Hessian of the log-marginal. This Hessian decomposes as
\begin{align}
    \nabla_{x_t}^2 \log p_t(x_t) &= \mathbb{E}\bigl[\nabla_{x_t}^2 \log p(x_t \mid x_1) \,\big|\, x_t\bigr] \nonumber \\ &\quad + \operatorname{Cov}\bigl(\nabla_{x_t}\log p(x_t \mid x_1) \,\big|\, x_t\bigr).
\label{eq:total_variance}
\end{align}
Both terms are available in closed form. The conditional score Eq.~\eqref{eq:cond_score} is affine in $x_t$, so its Hessian is the constant $-\frac{1}{(1{-}t)^2}\mathbf{I}$, unchanged under the posterior expectation. The only $x_1$-dependent part of the conditional score is $\frac{t}{(1{-}t)^2} x_1$, so its posterior covariance is $\frac{t^2}{(1{-}t)^4}\operatorname{Cov}(x_1 \mid x_t)$. Hence
\begin{equation}
\nabla_{x_t}^2 \log p_t(x_t) = -\frac{1}{(1{-}t)^2}\,\mathbf{I} + \frac{t^2}{(1{-}t)^4}\,\operatorname{Cov}(x_1 \mid x_t).
\label{eq:hessian_decomp}
\end{equation}
Substituting Eq.~\eqref{eq:hessian_decomp} into Eq.~\eqref{eq:jacobian_pmean} cancels the two constant terms:
\begin{align}
\nabla_{x_t}\mathbb{E}[x_1 \mid x_t]
&= \tfrac{1}{t}\mathbf{I} \nonumber \\
&\quad + \tfrac{(1{-}t)^2}{t}\!\left[-\tfrac{1}{(1{-}t)^2}\mathbf{I}
   + \tfrac{t^2}{(1{-}t)^4}\operatorname{Cov}(x_1 \mid x_t)\right] \nonumber \\
&= \tfrac{t}{(1{-}t)^2}\,\operatorname{Cov}(x_1 \mid x_t).
\label{eq:cancel}
\end{align}
Solving for the covariance gives the central identity:
\begin{equation}
\operatorname{Cov}(x_1 \mid x_t) = \frac{(1{-}t)^2}{t}\,\nabla_{x_t}\mathbb{E}[x_1 \mid x_t].
\label{eq:cov_from_jacobian}
\end{equation}
The posterior covariance equals the Jacobian of the posterior mean, scaled by $(1{-}t)^2/t$. It holds for any $p_1$ and any $t \in (0,1)$. The prefactor diverges as $t \to 0$, where $x_t$ is independent of $x_1$, and vanishes as $t \to 1$, where $x_t = x_1$.

\subsection{Velocity-Field Parameterization}
\label{sec:step3}

The posterior mean is determined by the velocity field. By Eq.~\eqref{eq:posterior_mean}, $\mathbb{E}[x_1 \mid x_t] = x_t + (1{-}t)\,v_\theta(x_t, t)$; differentiating gives
\begin{equation}
\nabla_{x_t}\mathbb{E}[x_1 \mid x_t] = \mathbf{I} + (1{-}t)\,J_{v_\theta}(x_t, t),
\label{eq:jac_substitute}
\end{equation}
with $J_{v_\theta} \coloneqq \nabla_{x_t}v_\theta \in \mathbb{R}^{d \times d}$. Substituting into Eq.~\eqref{eq:cov_from_jacobian} gives the working form
\begin{equation}
\operatorname{Cov}(x_1 \mid x_t) = \frac{(1{-}t)^2}{t}\bigl[\mathbf{I} + (1{-}t)\,J_{v_\theta}\bigr],
\label{eq:cov_final}
\end{equation}
and its trace,
\begin{equation}
U(x_t, t) \coloneqq \operatorname{Tr}\!\bigl(\operatorname{Cov}(x_1 \mid x_t)\bigr) = \frac{(1{-}t)^2}{t}\bigl[d + (1{-}t)\,\operatorname{div} v_\theta\bigr],
\label{eq:scalar_uq}
\end{equation}
where $\operatorname{div} v_\theta = \operatorname{Tr}(J_{v_\theta})$. The diagonal of Eq.~\eqref{eq:cov_final} gives per-pixel uncertainty and Eq.~\eqref{eq:scalar_uq} gives a scalar score, both from the pretrained velocity network.

The trace admits a direct reading. Along a trajectory, $\frac{d}{dt}\log p_t(x_t) = -\operatorname{div} v_\theta$, so $\operatorname{div} v_\theta$ is the local rate of volume change of the flow. A model transporting $\mathcal{N}(\mathbf{0},\mathbf{I})$ to a concentrated data distribution contracts volume on average, so $\operatorname{div} v_\theta < 0$ on average, and Eq.~\eqref{eq:scalar_uq} shifts the score below the baseline $\nicefrac{(1{-}t)^2}{t}\,d$ by $\frac{(1-t)^3}{t}\operatorname{div} v_\theta$. The diagonal of $J_{v_\theta}$ resolves this rate per coordinate.

\subsection{Stochastic Estimation}
\label{sec:computation}

Forming $J_{v_\theta}$ is infeasible in high dimensions ($d = 128^2$ for a brain MRI), but Eqs.~\eqref{eq:cov_final} and Eq.~\eqref{eq:scalar_uq} require only its diagonal and trace. The trace uses Hutchinson's estimator~\cite{hutchinson1989stochastic}: for Rademacher probes $\bm{\epsilon}$ drawn uniformly from $\{-1, +1\}^d$,
\begin{equation}
\operatorname{div} v_\theta = \operatorname{Tr}(J_{v_\theta}) = \mathbb{E}_{\bm{\epsilon}}\bigl[\bm{\epsilon}^\top J_{v_\theta}\,\bm{\epsilon}\bigr].
\label{eq:hutchinson}
\end{equation}
Each sample requires one Jacobian-vector product $J_{v_\theta}\bm{\epsilon}$, obtained in one forward-mode differentiation pass; with $S$ probes the cost is $S$ such products. Unless otherwise specified, we use $S=64$ probes. The diagonal follows from the same probes, $[J_{v_\theta}]_{ii} \approx \frac{1}{S}\sum_{s=1}^{S}\epsilon_i^{(s)}\,[J_{v_\theta}\bm{\epsilon}^{(s)}]_i$, and the per-pixel uncertainty is $\frac{(1-t)^2}{t}\bigl(1 + (1{-}t)\,[J_{v_\theta}]_{ii}\bigr)$.

\subsection{Single-Pass Uncertainty for One-Step Generators}
\label{sec:meanflow_uq}

Eq.~\eqref{eq:cov_from_jacobian} depends only on the posterior-mean map $x_t \mapsto \mathbb{E}[x_1 \mid x_t]$, whose Jacobian is fixed by the instantaneous velocity $v_\theta(x_t,t)$ through Eq.~\eqref{eq:posterior_mean}. A MeanFlow network supplies this velocity in one evaluation, so the identity gives the endpoint uncertainty without integration.

Let $\bar{u}_\theta(x,r,t)$ denote the average velocity over $[r,t]$; MeanFlow generates by $\hat{x}_1 = x_0 + \bar{u}_\theta(x_0,0,1)$. The average and instantaneous velocities satisfy
\begin{equation}
v_\theta(x_t,t) = \bar{u}_\theta(x_t,r,t) + (t-r)\,\tfrac{d}{dt}\bar{u}_\theta(x_t,r,t),
\label{eq:meanflow_identity}
\end{equation}
so $r=t$ recovers the instantaneous velocity from the same network, $v_\theta(x,s) = \bar{u}_\theta(x,s,s)$, which the MeanFlow objective fits on the diagonal $r=t$. Eq.~\eqref{eq:posterior_mean} requires this instantaneous velocity, not the interval average used for generation.\footnote{Using the generation map's Jacobian is not equivalent. Differentiating Eq.~\eqref{eq:meanflow_identity} gives $J_{v_\theta} = J_{\bar{u}_\theta} + (t-r)\,\nabla_x \tfrac{d}{dt}\bar{u}_\theta$, so substituting $J_{\bar{u}_\theta}$ into Eq.~\eqref{eq:cov_from_jacobian} leaves the term $\tfrac{(1-t)^2}{t}(1-t)(t-r)\,\nabla_x \tfrac{d}{dt}\bar{u}_\theta$. This term contains a second-order quantity and is $O(1)$ for one-step generation, where $t-r=1$. Evaluating at $r=t$ removes it, since $t-r=0$.}

Evaluating Eq.~\eqref{eq:cov_from_jacobian} at $t=\epsilon$ with $v_\theta(x_0,\epsilon)=\bar{u}_\theta(x_0,\epsilon,\epsilon)$ gives
\begin{equation}
    \operatorname{Cov}(x_1 \mid x_\epsilon)
    =
    \frac{(1-\epsilon)^2}{\epsilon}
    \bigl[
    \mathbf{I}
    +
    (1-\epsilon)\, J_{v_\theta}(x_0, \epsilon)
    \bigr],
    \label{eq:meanflow_cov}
\end{equation}
with $J_{v_\theta}(x_0,\epsilon) = \nabla_x v_\theta(x,\epsilon)\big|_{x=x_0}$; we set $\epsilon = 10^{-2}$. Since $x_\epsilon = x_0 + O(\epsilon)$ is the generator input, Eq.~\eqref{eq:meanflow_cov} is the covariance of the endpoint that $x_0$ maps to, computed from one evaluation of $v_\theta(\cdot,\epsilon)$ and one Jacobian-vector product per probe, with no integration.

Two conditions fix the estimand. As $\epsilon \to 0$ the prefactor $(1-\epsilon)^2/\epsilon$ diverges; at the population optimum the bracket vanishes at the same rate and the product tends to the marginal covariance $\operatorname{Cov}(x_1)$, so we fix $\epsilon > 0$ and evaluate on the trained network. For a trained network $J_{v_\theta}$ need not be symmetric, so the diagonal entries $1 + (1{-}t)\,[J_{v_\theta}]_{ii}$ and the trace can be negative at small $t$; we report both and set $U \leftarrow \max(U, 0)$.

\section{Experiments}
We assess the proposed closed-form uncertainty quantification (UQ) method on two benchmark datasets of different scale and visual complexity: CIFAR-10~\cite{krizhevsky2009learning} and BraTS~\cite{menze2014multimodal}. The experiments demonstrate the method’s effectiveness from four complementary perspectives. i. Fidelity. We measure the concordance between the proposed estimator and the true reconstruction error across interpolation times, at both pixel and sample levels, and compare it with established UQ baselines. ii. Interpretability. We assess the spatial coherence, anatomical plausibility, and semantic informativeness of the pixel-wise uncertainty maps. iii. Robustness. We analyze the estimator’s sensitivity to the number of Hutchinson probes and identify the minimum number needed for stable uncertainty maps. iv. Efficiency. We compare the computational cost of the proposed estimator with existing UQ baselines.

\subsection{Setup}
\label{sec:expt_setup}

\paragraph{Data.}

On CIFAR-10 ($32\times32$), we train an unconditional MeanFlow model from scratch using a SongUNet backbone with positional time embeddings. 
The model follows the DDPM++-style architecture and is trained on RGB images with the MeanFlow objective.
For BraTS, we train a separate unconditional MeanFlow model on preprocessed 2D axial MRI slices. 
Each slice is resized to $128\times128$ and represented as a single-channel grayscale image. 
To accommodate the higher resolution and the smaller medical dataset, we use a lighter SongUNet backbone with a reduced base channel width, disable data augmentation, and train with Adam using a smaller learning rate. 
For uncertainty baselines, all methods are evaluated using the corresponding dataset-specific trained backbone, ensuring that differences arise from the uncertainty estimation procedure rather than from the generative model architecture.

\begin{table*}[t] 
\centering

\setlength{\tabcolsep}{3pt} 
\resizebox{\textwidth}{!}{ 
\begin{tabular}{ccccccccc} 
\toprule 
\textbf{Metric} & \textbf{Method} 
& $\boldsymbol{t=0.3}$ & $\boldsymbol{t=0.4}$ & $\boldsymbol{t=0.5}$ 
& $\boldsymbol{t=0.6}$ & $\boldsymbol{t=0.7}$ & $\boldsymbol{t=0.8}$ & $\boldsymbol{t=0.9}$ \\ 
\midrule 

\multirow{5}{*}{$\rho_{\rm pix} \uparrow$} 
& Ensemble & $0.223 \pm 0.006$ & $0.269 \pm 0.002$ & $0.293 \pm 0.003$ & $0.303 \pm 0.002$ & $0.296 \pm 0.004$ & $0.276 \pm 0.003$ & $0.225 \pm 0.001$ \\ 
& MCDrop & $0.265 \pm 0.003$ & $0.320 \pm 0.003$ & $0.354 \pm 0.003$ & $0.373 \pm 0.002$ & $0.373 \pm 0.003$ & $0.352 \pm 0.002$ & $0.285 \pm 0.002$ \\ 
& BayesDiff & $0.070 \pm 0.006$ & $0.099 \pm 0.006$ & $0.123 \pm 0.006$ & $0.142 \pm 0.006$ & $0.163 \pm 0.005$ & $0.181 \pm 0.003$ & $0.168 \pm 0.002$ \\ 
& UA-Flow & $\textbf{0.272} \pm \textbf{0.004}$ & $\textbf{0.322} \pm \textbf{0.004}$ & $0.347 \pm 0.004$ & $0.355 \pm 0.003$ & $0.346 \pm 0.003$ & $0.326 \pm 0.003$ & $0.285 \pm 0.003$ \\ 
& Ours & $0.242 \pm 0.003$ & $0.311 \pm 0.002$ & $\textbf{0.360} \pm \textbf{0.002}$ & $\textbf{0.392} \pm \textbf{0.003}$ & $\textbf{0.407} \pm \textbf{0.003}$ & $\textbf{0.405} \pm \textbf{0.001}$ & $\textbf{0.367} \pm \textbf{0.002}$ \\ 
\midrule 

\multirow{5}{*}{$\mathrm{hit}@30\% \uparrow$} 
& Ensemble & $0.411 \pm 0.005$ & $0.433 \pm 0.003$ & $0.442 \pm 0.002$ & $0.445 \pm 0.002$ & $0.440 \pm 0.002$ & $0.427 \pm 0.003$ & $0.402 \pm 0.001$ \\ 
& MCDrop & $0.431 \pm 0.003$ & $0.457 \pm 0.003$ & $0.474 \pm 0.002$ & $0.481 \pm 0.002$ & $0.479 \pm 0.001$ & $0.465 \pm 0.002$ & $0.430 \pm 0.002$ \\ 
& BayesDiff & $0.335 \pm 0.004$ & $0.350 \pm 0.005$ & $0.363 \pm 0.004$ & $0.372 \pm 0.005$ & $0.381 \pm 0.005$ & $0.386 \pm 0.004$ & $0.378 \pm 0.002$ \\ 
& UA-Flow & $\textbf{0.443} \pm \textbf{0.002}$ & $\textbf{0.466} \pm \textbf{0.002}$ & $0.475 \pm 0.003$ & $0.474 \pm 0.003$ & $0.464 \pm 0.002$ & $0.449 \pm 0.001$ & $0.427 \pm 0.001$ \\ 
& Ours & $0.429 \pm 0.002$ & $0.461 \pm 0.003$ & $\textbf{0.484} \pm \textbf{0.002}$ & $\textbf{0.497} \pm \textbf{0.003}$ & $\textbf{0.500} \pm \textbf{0.002}$ & $\textbf{0.493} \pm \textbf{0.002}$ & $\textbf{0.467} \pm \textbf{0.001}$ \\ 
\midrule 

\multirow{5}{*}{$\rho_{\rm samp} \uparrow$} 
& Ensemble & $0.377 \pm 0.048$ & $0.404 \pm 0.035$ & $0.348 \pm 0.043$ & $0.329 \pm 0.042$ & $0.343 \pm 0.035$ & $0.399 \pm 0.014$ & $0.415 \pm 0.008$ \\ 
& MCDrop & $0.424 \pm 0.059$ & $0.469 \pm 0.049$ & $0.456 \pm 0.023$ & $0.476 \pm 0.024$ & $0.483 \pm 0.013$ & $0.499 \pm 0.018$ & $0.493 \pm 0.007$ \\ 
& BayesDiff & $0.379 \pm 0.098$ & $0.385 \pm 0.063$ & $0.396 \pm 0.037$ & $0.408 \pm 0.029$ & $0.412 \pm 0.022$ & $0.404 \pm 0.016$ & $0.283 \pm 0.007$ \\ 
& UA-Flow & $0.562 \pm 0.049$ & $0.654 \pm 0.043$ & $0.733 \pm 0.022$ & $0.800 \pm 0.010$ & $0.839 \pm 0.008$ & $0.858 \pm 0.006$ & $0.833 \pm 0.004$ \\ 
& Ours & $\textbf{0.679} \pm \textbf{0.055}$ & $\textbf{0.791} \pm \textbf{0.042}$ & $\textbf{0.856} \pm \textbf{0.017}$ & $\textbf{0.906} \pm \textbf{0.006}$ & $\textbf{0.938} \pm \textbf{0.006}$ & $\textbf{0.956} \pm \textbf{0.005}$ & $\textbf{0.966} \pm \textbf{0.001}$ \\ 
\bottomrule 
\end{tabular}} 
\caption{Uncertainty--error consistency on the CIFAR-10 dataset at different interpolation times. Results are reported as mean $\pm$ std over 5 repeated runs. The best result in each group is
shown in \textbf{bold}.} 
\label{tab:uncertainty_error_all}
\end{table*}

\paragraph{Experiment Setup.} We compare four UQ baselines.
\textbf{Ensemble}~\cite{lakshminarayanan2017simple} computes the variance of $\mathbb{E}[x_1|x_t]$ across the 5 independently trained MeanFlow models.
\textbf{MC~Dropout} (MCDrop)~\cite{gal2016dropout} computes the variance across 500 stochastic forward passes through the dropout-enabled model.
\textbf{BayesDiff-style} (BayesDiff)~\cite{kou2024bayesdiff} is adapted as a last-layer Laplace baseline, where we fit a diagonal posterior over the final output layer and use the resulting predictive variance as the uncertainty estimate.
\textbf{UA-Flow-style} (UA-Flow)~\cite{han2026uaflow} trains a lightweight variance head on top of the frozen MeanFlow model to predict heteroscedastic velocity uncertainty from the corrected velocity residual.

\begin{figure}[h]
    \centering
    \includegraphics[width=\linewidth]{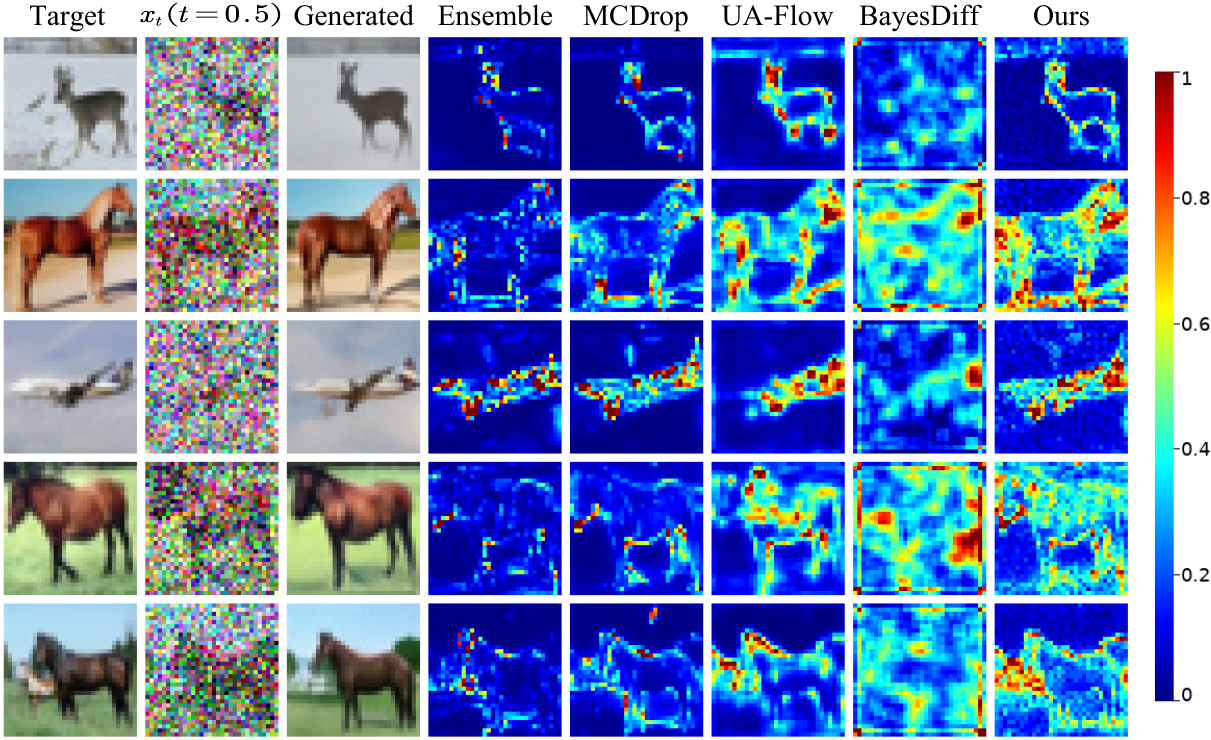}
    \caption{Per-pixel uncertainty maps on CIFAR-10 ($32 \times 32$) at the intermediate state $x_t$ with $t=0.5$.}
    \label{fig:visualization}
\end{figure}
\subsection{Fidelity \& Interpretability}

\paragraph{Uncertainty Visualization.} Fig.~\ref{fig:visualization} shows representative CIFAR-10 qualitative results at $t = 0.5$. Each row contains the target image, the noisy intermediate state $x_t$, the MeanFlow reconstruction, and uncertainty maps from Ensemble, MCDrop, UA-Flow, BayesDiff, and our method, all normalized to a common color scale. 
The reconstructions mostly recover the global object layout, but fine structures (e.g., contours, legs, heads, tails, thin boundaries) remain difficult to reconstruct. Our method assigns higher uncertainty to these ambiguous regions, producing maps that align well with object silhouettes and reconstruction-sensitive details. Compared with Ensemble and MCDrop, it yields more spatially coherent and continuous maps, rather than sparse, isolated responses. 
UA-Flow sometimes highlights boundaries but often diffuses uncertainty into uninformative background areas. BayesDiff produces smoother but less localized maps, reducing its ability to pinpoint specific ambiguous structures. 
In contrast, our closed-form estimator consistently emphasizes object boundaries and fine-scale structures, improving the interpretability of spatial uncertainty patterns.

\paragraph{Uncertainty-Error Consistency.}
\label{sec:ue_consistency}
Tab.~\ref{tab:uncertainty_error_all} evaluates the alignment between uncertainty estimates and reconstruction errors on CIFAR-10 across interpolation times, measured by pixel-level Spearman correlation ($\rho_{\rm pix}$), the top-30\% uncertainty–error hit rate (hit@$30\%$), and sample-level Spearman correlation ($\rho_{\rm samp}$). Our method yields the strongest overall alignment, particularly at middle and late stages. While UA-Flow excels at early noisy states ($t=0.3, 0.4$), our approach dominates both $\rho_{\rm pix}$ and hit@$30\%$ from $t=0.5$ to $t=0.9$, peaking at $\rho_{\rm pix}=0.405$ ($t=0.8$). The most significant gains occur at the sample level: ours consistently achieves the highest $\rho_{\rm samp}$ across all times, increasing from $0.679$ at $t=0.3$ to $0.966$ at $t=0.9$. Conversely, baselines like Ensemble, MCDrop, and BayesDiff exhibit substantially weaker sample-level correlations. Overall, our method not only faithfully matches local reconstruction-error structures but also provides a more reliable global ranking of sample difficulty than sampling- or retraining-based baselines.
\begin{figure*}[h]
    \centering
    \includegraphics[width=\linewidth]{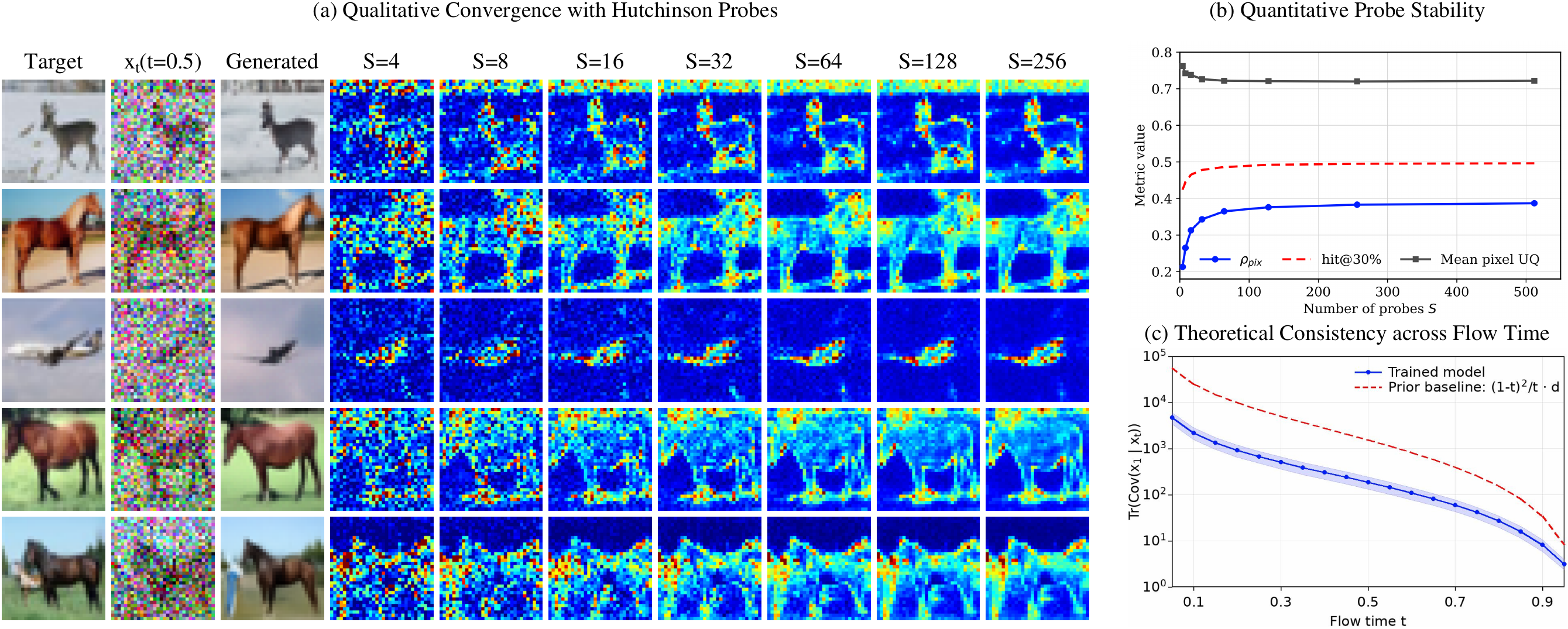}
    \caption{
Efficiency, stability, and empirical validation of MeanFlow posterior uncertainty.
}
    \label{fig:hutchinson_probes}
\end{figure*}

\paragraph{Geometric Boundary Consistency}
\label{sec:geometric_consistency}

Tab.~\ref{tab:geometric_boundary} reports the geometric consistency between uncertainty and reconstruction-error maps using the boundary F1 score (Boundary F1), average symmetric surface distance (ASSD), and 95th-percentile Hausdorff distance (HD95). Across all Top-K ratios, our method consistently achieves the best performance. This advantage is most evident under the strict Top-$10\%$ setting, where ours yields a Boundary F1 of 0.692 and reduces ASSD and HD95 to 0.033 and 0.102, respectively. This demonstrates our method's ability to concentrate uncertainty on the most relevant error regions even with a small pixel fraction. As K increases to $20\%$ and $30\%$, our approach remains superior, outperforming the strong ensemble baseline even at Top-$30\%$ (Boundary F1: 0.857, ASSD: 0.020, HD95: 0.068). Overall, these results show that our closed-form estimate more faithfully captures the geometry of high-error regions than the compared baselines.

\begin{table}[tbh]
\centering

\setlength{\tabcolsep}{3pt}
\begin{tabular}{ccccc}
\toprule
\textbf{Top-$K$}
& \textbf{Method}
& \textbf{Boundary F1} $\uparrow$
& \textbf{ASSD} $\downarrow$
& \textbf{HD95} $\downarrow$ \\
\midrule

\multirow{5}{*}{10\%}
& Ensemble & 0.596 & 0.042 & 0.125 \\
& MCDrop & 0.620 & 0.040 & 0.128 \\
& BayesDiff  & 0.439 & 0.061 & 0.170 \\
& UA-Flow & 0.632 & 0.039 & 0.131 \\
& Ours & \textbf{0.692} & \textbf{0.033} & \textbf{0.102} \\

\midrule

\multirow{5}{*}{20\%}
& Ensemble & 0.758 & 0.028 & 0.089 \\
& MCDrop & 0.759 & 0.028 & 0.097 \\
& BayesDiff  & 0.622 & 0.039 & 0.118 \\
& UA-Flow & 0.745 & 0.029 & 0.106 \\
& Ours & \textbf{0.809} & \textbf{0.024} & \textbf{0.079} \\

\midrule

\multirow{5}{*}{30\%}
& Ensemble & 0.835 & 0.022 & 0.072 \\
& MCDrop & 0.811 & 0.023 & 0.084 \\
& BayesDiff  & 0.735 & 0.029 & 0.090 \\
& UA-Flow & 0.790 & 0.025 & 0.092 \\
& Ours & \textbf{0.857} & \textbf{0.020} & \textbf{0.068} \\

\bottomrule
\end{tabular}

\caption{Geometric boundary consistency. ASSD and
HD95 are normalized by the image diagonal. The best result in each group is
shown in \textbf{bold}.}
\label{tab:geometric_boundary}
\end{table}

\subsection{Computational Stability \& Efficiency}
\paragraph{Probe efficiency and stability.}
\label{sec:probe_stability}
Fig.~\ref{fig:hutchinson_probes}(a) visualizes the MeanFlow posterior uncertainty across varying Hutchinson probe counts $S$. At low probe counts, the estimates exhibit high variance and unstructured noise. As $S$ increases, the uncertainty maps smoothly converge, aligning precisely with semantic boundaries and ambiguous regions. Visual changes become negligible beyond $S=64$, indicating spatial pattern convergence. Fig.~\ref{fig:hutchinson_probes}(b) quantitatively confirms this trend: both $\rho_{\rm pix}$ and hit@30\% plateau after $S=64$, while the mean pixel uncertainty stabilizes. Increasing the probe count effectively reduces estimator variance without altering the underlying uncertainty signal. We adopt $S=64$ to optimally balance estimation accuracy and computational cost.

\paragraph{Empirical signature across flow time.}
Fig.~\ref{fig:hutchinson_probes}(c) compares the scalar uncertainty $U(x_t,t)=\mathrm{Tr}(\mathrm{Cov}(x_1\mid x_t))$ against the prior baseline $\nicefrac{(1-t)^2}{t} \cdot d$. The empirical uncertainty decreases monotonically as $t \to 1$ and remains strictly bounded below the baseline. This gap confirms a negative divergence contribution from the learned velocity field, corroborating the contractive posterior covariance predicted by Eq.~\eqref{eq:scalar_uq}. Furthermore, the narrow variability band demonstrates robust estimator stability across the entire flow trajectory.

\begin{table}[h]
\centering

\small
\setlength{\tabcolsep}{2.0pt}
\begin{tabular}{@{}cccccc@{}}
\toprule
Method & Training-free & Net-Agn & Offline (h) & Online (s) & Total \\
\midrule
MCDrop    & \checkmark & $\times$   & --    & 8.368 & 8.368 s \\
BayesDiff & $\times$   & $\times$   & 3.32  & 0.018 & $\sim$3.32 h \\
UA-Flow   & $\times$   & $\times$   & 15.17 & 0.018 & $\sim$15.17 h \\
Ours & \textbf{\checkmark} & \textbf{\checkmark} & \textbf{--} & 9.204 & 9.204 s \\
\bottomrule
\end{tabular}
\caption{Computational efficiency on identical hardware. Training-free and Net-Agn indicate no extra method-specific fitting and network agnosticism, respectively. Online time is averaged over 10 runs; Total is Offline plus Online for one sample. “--” denotes no offline cost. Our method matches MCDrop in speed while providing UA-Flow–level (or better) uncertainty, without UA-Flow’s long offline training.}
\label{tab:efficiency}
\end{table}

Tab.~\ref{tab:efficiency} compares the computational cost and deployment flexibility of various UQ methods. BayesDiff and UA-Flow require substantial offline training ($\sim$3.32 and 15.17 hours, respectively). While they offer low online latency, this initial setup limits their immediate deployment. In contrast, MCDrop and our method are training-free and can be applied directly. However, MCDrop requires specific architectural features (i.e., built-in dropout layers) and, as shown previously, yields less accurate uncertainty estimates. Our approach is network-agnostic and operates post hoc on standard pretrained models. By avoiding offline training costs and architectural constraints, our method improves estimation accuracy and offers practical end-to-end efficiency, with a reasonable online cost of 9.204 seconds per sample.

\subsection{Clinical Utility on Brain MRI}

\begin{figure}[!tbh]
    \centering
    \includegraphics[width=\linewidth]{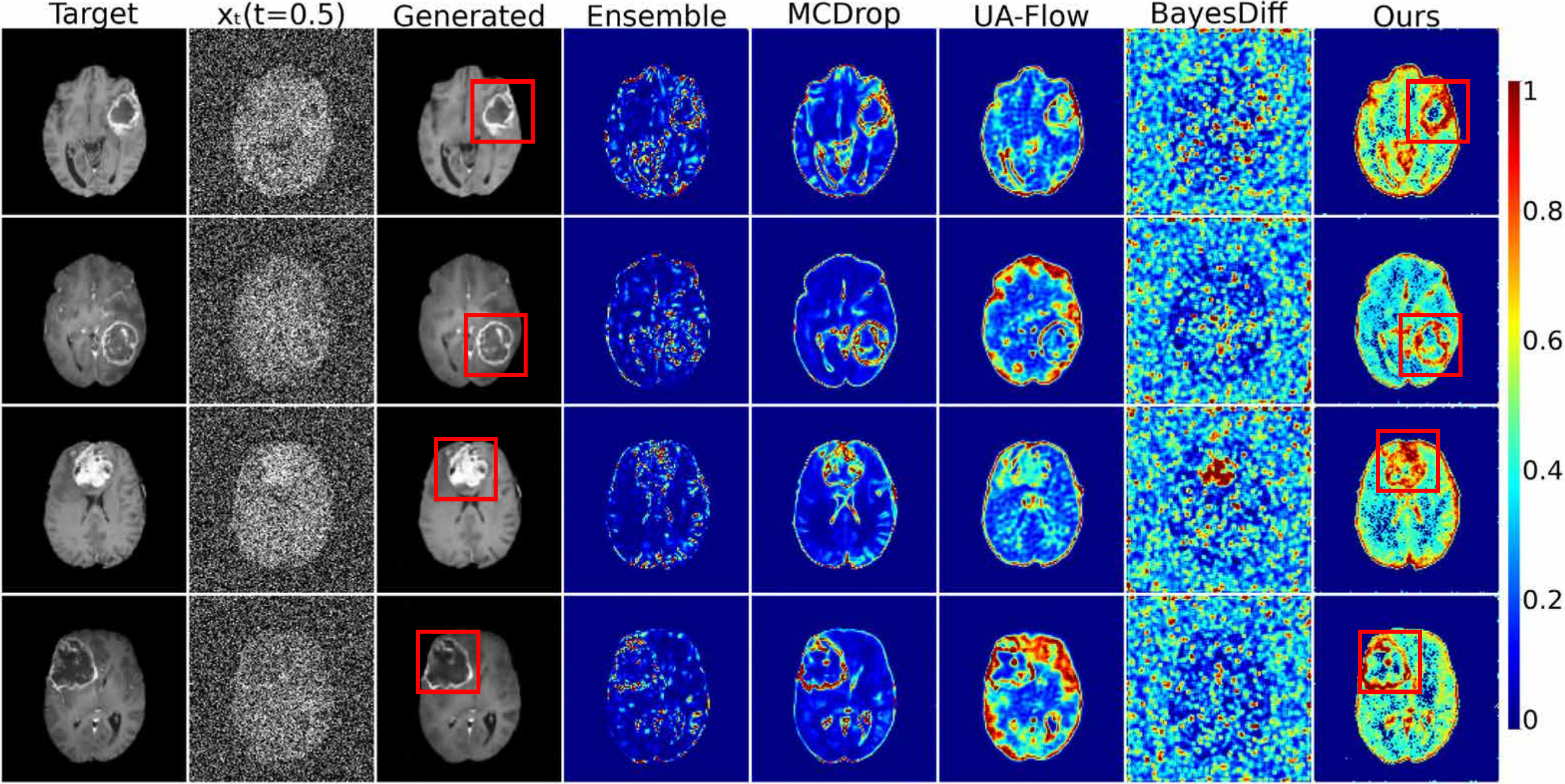}
    \caption{Anatomical uncertainty localization on BraTS.}
    \label{fig:braTS}
\end{figure}

Fig.~\ref{fig:braTS} visualizes anatomical uncertainty localization on the BraTS dataset at $t=0.5$. Red boxes denote reconstruction-sensitive tumor regions and adjacent anatomical interfaces across the target MRI, noisy state $x_t$, MeanFlow reconstruction, and corresponding uncertainty maps. Both Ensemble and MCDrop yield diffuse responses, struggling to localize the heterogeneous tumor core, whereas BayesDiff is corrupted by high-frequency background noise. Although UA-Flow captures coarse lesion structures, its uncertainty estimates bleed broadly into surrounding healthy tissues. In contrast, our approach delivers structurally crisp maps that precisely delineate ambiguous lesion boundaries and internal tumor morphology without introducing distracting artifacts. Our closed-form estimator reliably identifies anatomically meaningful uncertainty in medical images, completely obviating the need for stochastic sampling or auxiliary networks.

Tab.~\ref{tab:tumor_boundary_braTS} reports the geometric correspondence between the top-$K$ most uncertain intra-tumoral regions and the ground-truth lesion contours. At Top-$10\%$, the proposed method achieves the lowest HD95 (7.383) and remains close to MCDrop in Boundary F1 (0.672 vs.\ 0.674) and ASSD (2.073 vs.\ 2.027). At Top-$20\%$, it outperforms all competitors on all three metrics, with a Boundary F1 of 0.681, an ASSD of 1.841, and an HD95 of 8.258. At Top-$30\%$, it attains the highest Boundary F1 (0.647), with slightly worse ASSD (1.868) and HD95 (8.766) than MCDrop.

\begin{table}[tbh]
\centering

\setlength{\tabcolsep}{3pt}

\begin{tabular}{ccccc}
\toprule
\textbf{Top-$K$}
& \textbf{Method}
& \textbf{Boundary F1} $\uparrow$
& \textbf{ASSD} $\downarrow$
& \textbf{HD95} $\downarrow$ \\
\midrule

\multirow{5}{*}{10\%}
& Ensemble
& $0.620$
& $2.277$
& $8.077$ \\

& MCDrop
& \textbf{0.674}
& \textbf{2.027}
& $7.508$ \\

& BayesDiff
& $0.466$
& $3.014$
& $8.492$ \\

& UA-Flow
& $0.420$
& $3.350$
& $8.942$ \\

& Ours
& $0.672$
& $2.073$
& \textbf{7.383} \\
\midrule

\multirow{5}{*}{20\%}
& Ensemble
& $0.601$
& $2.177$
& $8.773$ \\

& MCDrop
& $0.676$
& $1.849$
& $8.272$ \\

& BayesDiff
& $0.471$
& $2.812$
& $9.143$ \\

& UA-Flow
& $0.423$
& $2.993$
& $9.301$ \\

& Ours
& \textbf{0.681}
& \textbf{1.841}
& \textbf{8.258} \\
\midrule

\multirow{5}{*}{30\%}
& Ensemble
& $0.562$
& $2.200$
& $9.077$ \\

& MCDrop
& $0.645$
& \textbf{1.851}
& \textbf{8.731} \\

& BayesDiff
& $0.460$
& $2.704$
& $9.367$ \\

& UA-Flow
& $0.421$
& $2.823$
& $9.385$ \\

& Ours
& \textbf{0.647}
& $1.868$
& $8.766$ \\
\bottomrule
\end{tabular}
\caption{Brain tumor boundary consistency across different uncertainty estimation methods. The best result in each group is shown in \textbf{bold}.}
\label{tab:tumor_boundary_braTS}
\end{table}

\section{Conclusion}
\label{sec:conclusion}

In this work, we establish a theoretically rigorous and computationally elegant framework for uncertainty quantification in flow-matching models. By deriving an exact, closed-form posterior covariance rooted directly in the Jacobian of the learned velocity field, we bridge a critical gap between generative performance and predictive reliability. Our formulation operates entirely post hoc and obviates the need for retraining, architectural modifications, or auxiliary modules. By reducing the scalar uncertainty to the velocity field's divergence, we unlock highly scalable estimation via Hutchinson's trace estimator, effectively characterizing the complete generative uncertainty for one-step models like MeanFlow. Beyond strong empirical alignment with reconstruction errors on CIFAR-10, our method demonstrates profound practical value in safety-critical contexts. When applied to brain MRI with real tumors, it precisely localizes complex, clinically relevant anatomical ambiguities without hallucinating structural artifacts. By completely eliminating offline training overhead while preserving online efficiency, this work elevates flow matching from a purely generative engine to a robust, interpretable, and network-agnostic framework, paving the way for its safe deployment in risk-sensitive computer vision tasks and advanced medical imaging applications.

\bibliography{main}

\appendix
\section*{Appendix}
\setcounter{section}{0}
\setcounter{equation}{0}
\renewcommand{\theequation}{A.\arabic{equation}}
\setcounter{figure}{0}
\renewcommand{\thefigure}{A\arabic{figure}}
\setcounter{table}{0}
\renewcommand{\thetable}{A\arabic{table}}

\section{Derivation of the Posterior Covariance Identity}
\label{sec:posterior_covariance_derivation}
In this section, we provide a complete derivation of the posterior
covariance identity for the linear Flow Matching interpolant. We first
express the posterior mean of the target sample in terms of the marginal
score using Fisher's identity and a Tweedie-type formula. We then
differentiate this posterior mean and apply the conditional Hessian
decomposition to relate its Jacobian to the posterior covariance. This
yields a closed-form identity that expresses
\(\Cov(x_1\mid x_t)\) through the local Jacobian of the posterior-mean
map, which can subsequently be connected to the Jacobian of the
population-optimal Flow Matching velocity field.
\subsection{Setup and Notation}

Let \(x_1\in\R^d\) be a data sample drawn from the target distribution
\(p_1\), and let \(x_0\in\R^d\) be an independent standard Gaussian sample:
\begin{equation}
  x_1\sim p_1,
  \qquad
  x_0\sim\mathcal N(0,I),
  \qquad
  x_0\perp x_1.
\end{equation}
For \(t\in(0,1)\), consider the linear Flow Matching interpolant
\begin{equation}
  x_t=t x_1+(1-t)x_0.
  \label{app:eq:interpolant}
\end{equation}
We denote the marginal density of \(x_t\) by \(p_t(x_t)\) and define
the posterior mean
\begin{equation}
  m_t(x_t):=\E[x_1\mid x_t].
  \label{eq:posterior_mean_definition}
\end{equation}
All gradients and Jacobians below are taken with respect to \(x_t\),
with \(t\) held fixed. For a vector-valued function
\(f:\R^d\rightarrow\R^d\), we use the Jacobian convention
\begin{equation}
  [\nabla_{x_t}f(x_t)]_{ij}
  =
  \frac{\partial f_i(x_t)}{\partial x_{t,j}}.
\end{equation}
We assume the regularity conditions required to differentiate under the
integral sign.

Conditioned on \(x_1\), the interpolant is Gaussian:
\begin{equation}
  x_t\mid x_1
  \sim
  \mathcal N\!\left(t x_1,(1-t)^2I\right).
  \label{eq:conditional_distribution}
\end{equation}
Its conditional score is therefore
\begin{equation}
  \nabla_{x_t}\log p(x_t\mid x_1)
  =
  \frac{t x_1-x_t}{(1-t)^2}.
  \label{eq:conditional_score}
\end{equation}

\subsection{Posterior Mean Identity}

The marginal density of \(x_t\) is
\begin{equation}
  p_t(x_t)
  =
  \int p(x_t\mid x_1)p_1(x_1)\,dx_1.
\end{equation}
Differentiating under the integral sign gives Fisher's identity:
\begin{align}
  \nabla_{x_t}\log p_t(x_t)
  &=
  \frac{
    \int
    \nabla_{x_t}p(x_t\mid x_1)p_1(x_1)\,dx_1
  }{
    p_t(x_t)
  } \\
  &=
  \E\!\left[
    \nabla_{x_t}\log p(x_t\mid x_1)
    \mid x_t
  \right].
  \label{eq:fisher_identity}
\end{align}
Substituting \eqref{eq:conditional_score} into
\eqref{eq:fisher_identity} yields
\begin{align}
  \nabla_{x_t}\log p_t(x_t)
  &=
  \E\!\left[
    \frac{t x_1-x_t}{(1-t)^2}
    \,\middle|\,x_t
  \right] \\
  &=
  \frac{t\,m_t(x_t)-x_t}{(1-t)^2}.
  \label{eq:marginal_score_mean}
\end{align}
Solving for \(m_t(x_t)\), we obtain the posterior mean identity
\begin{equation}
  \E[x_1\mid x_t]
  =
  \frac{x_t}{t}
  +
  \frac{(1-t)^2}{t}
  \nabla_{x_t}\log p_t(x_t).
  \label{eq:posterior_mean_identity}
\end{equation}
This is the Tweedie identity associated with the linear Flow Matching
interpolant.

The population-optimal Flow Matching velocity is
\begin{equation}
  v^*(x_t,t)
  :=
  \E[x_1-x_0\mid x_t].
  \label{eq:optimal_velocity}
\end{equation}
From \eqref{app:eq:interpolant},
\begin{equation}
  x_1-x_0
  =
  \frac{x_1-x_t}{1-t}.
\end{equation}
Taking the conditional expectation given \(x_t\) gives
\begin{equation}
  v^*(x_t,t)
  =
  \frac{m_t(x_t)-x_t}{1-t}.
\end{equation}
Hence, the posterior mean can equivalently be written as
\begin{equation}
  m_t(x_t)
  =
  x_t+(1-t)v^*(x_t,t).
  \label{eq:posterior_mean_velocity}
\end{equation}

\subsection{Posterior Covariance Identity}
\label{sec:posterior_covariance_identity}

Differentiating the posterior mean identity
\eqref{eq:posterior_mean_identity} with respect to \(x_t\) gives
\begin{equation}
  \nabla_{x_t}\E[x_1\mid x_t]
  =
  \frac{1}{t}I
  +
  \frac{(1-t)^2}{t}
  \nabla_{x_t}^2\log p_t(x_t).
  \label{app:eq:jacobian_pmean}
\end{equation}
The marginal Hessian admits the decomposition
\begin{multline}
  \nabla_{x_t}^2\log p_t(x_t)
  =
  \E\!\left[
    \nabla_{x_t}^2\log p(x_t\mid x_1)
    \mid x_t
  \right] \\
  +
  \Cov\!\left(
    \nabla_{x_t}\log p(x_t\mid x_1)
    \mid x_t
  \right).
  \label{eq:hessian_decomposition}
\end{multline}
Using the conditional score in \eqref{eq:conditional_score}, the two terms are
\begin{equation}
  \E\!\left[
    \nabla_{x_t}^2\log p(x_t\mid x_1)
    \mid x_t
  \right]
  =
  -\frac{1}{(1-t)^2}I
\end{equation}
and
\begin{equation}
  \Cov\!\left(
    \nabla_{x_t}\log p(x_t\mid x_1)
    \mid x_t
  \right)
  =
  \frac{t^2}{(1-t)^4}
  \Cov(x_1\mid x_t).
\end{equation}
Therefore,
\begin{equation}
  \nabla_{x_t}^2\log p_t(x_t)
  =
  -\frac{1}{(1-t)^2}I
  +
  \frac{t^2}{(1-t)^4}
  \Cov(x_1\mid x_t).
  \label{eq:hessian_covariance}
\end{equation}
Substituting this result into \eqref{app:eq:jacobian_pmean}, the two
identity-matrix terms cancel, yielding
\begin{equation}
  \nabla_{x_t}\E[x_1\mid x_t]
  =
  \frac{t}{(1-t)^2}\Cov(x_1\mid x_t).
\end{equation}
Hence,
\begin{equation}
  \Cov(x_1\mid x_t)
  =
  \frac{(1-t)^2}{t}
  \nabla_{x_t}\E[x_1\mid x_t].
  \label{app:eq:cov_from_jacobian}
\end{equation}

\section{Additional Qualitative Results}

\begin{figure}[t]
    \centering
    \includegraphics[width=\linewidth]{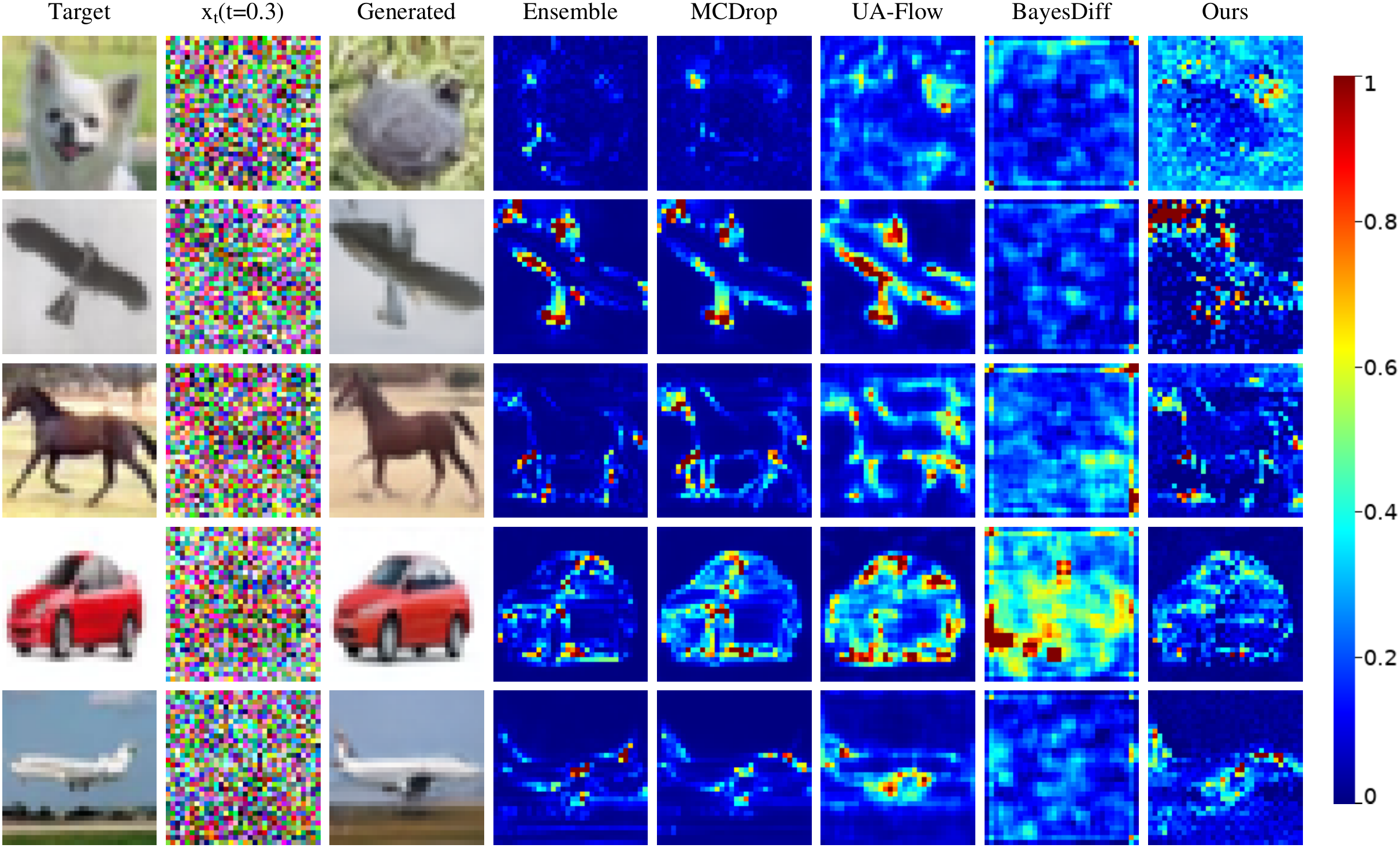}
    \caption{Per-pixel uncertainty maps on CIFAR-10 at interpolation
    time $t=0.3$.}
    \label{fig:cifar-sup-t03}
\end{figure}

\begin{figure}[t]
    \centering
    \includegraphics[width=\linewidth]{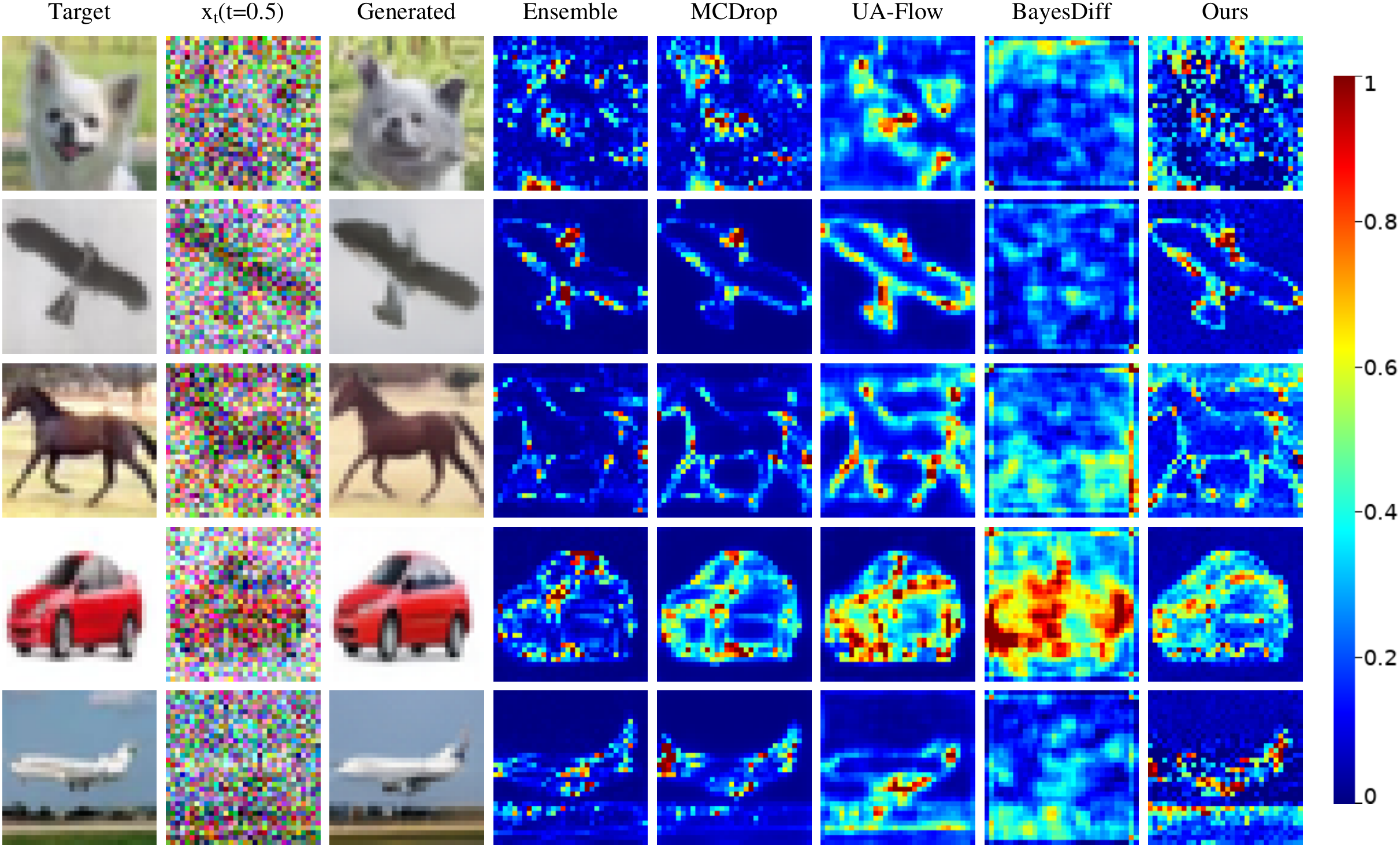}
    \caption{Per-pixel uncertainty maps on CIFAR-10 at interpolation
    time $t=0.5$.}
    \label{fig:cifar-sup-t05}
\end{figure}

\begin{figure}[t]
    \centering
    \includegraphics[width=\linewidth]{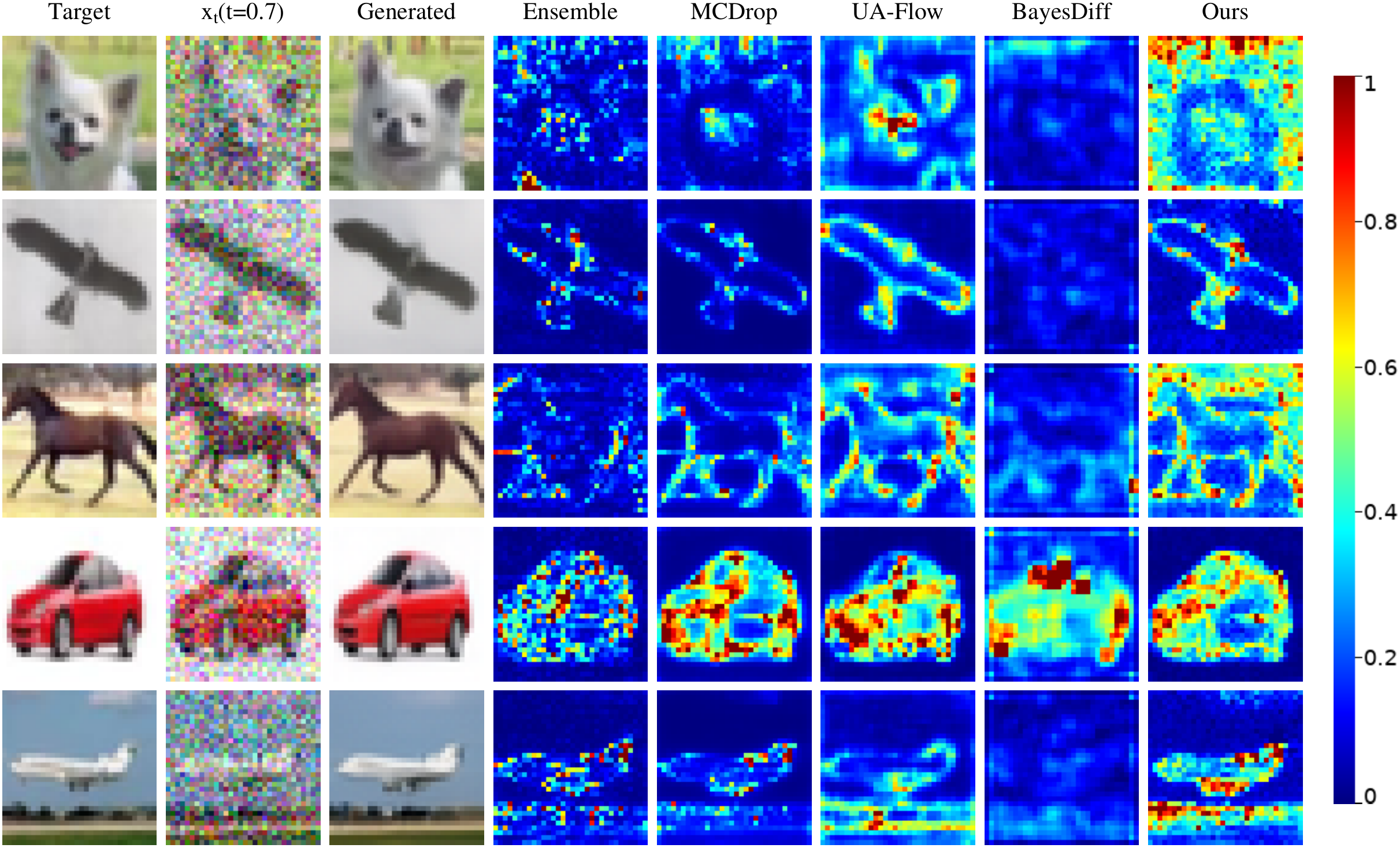}
    \caption{Per-pixel uncertainty maps on CIFAR-10 at interpolation
    time $t=0.7$.}
    \label{fig:cifar-sup-t07}
\end{figure}

\begin{figure}[t]
    \centering
    \includegraphics[width=\linewidth]{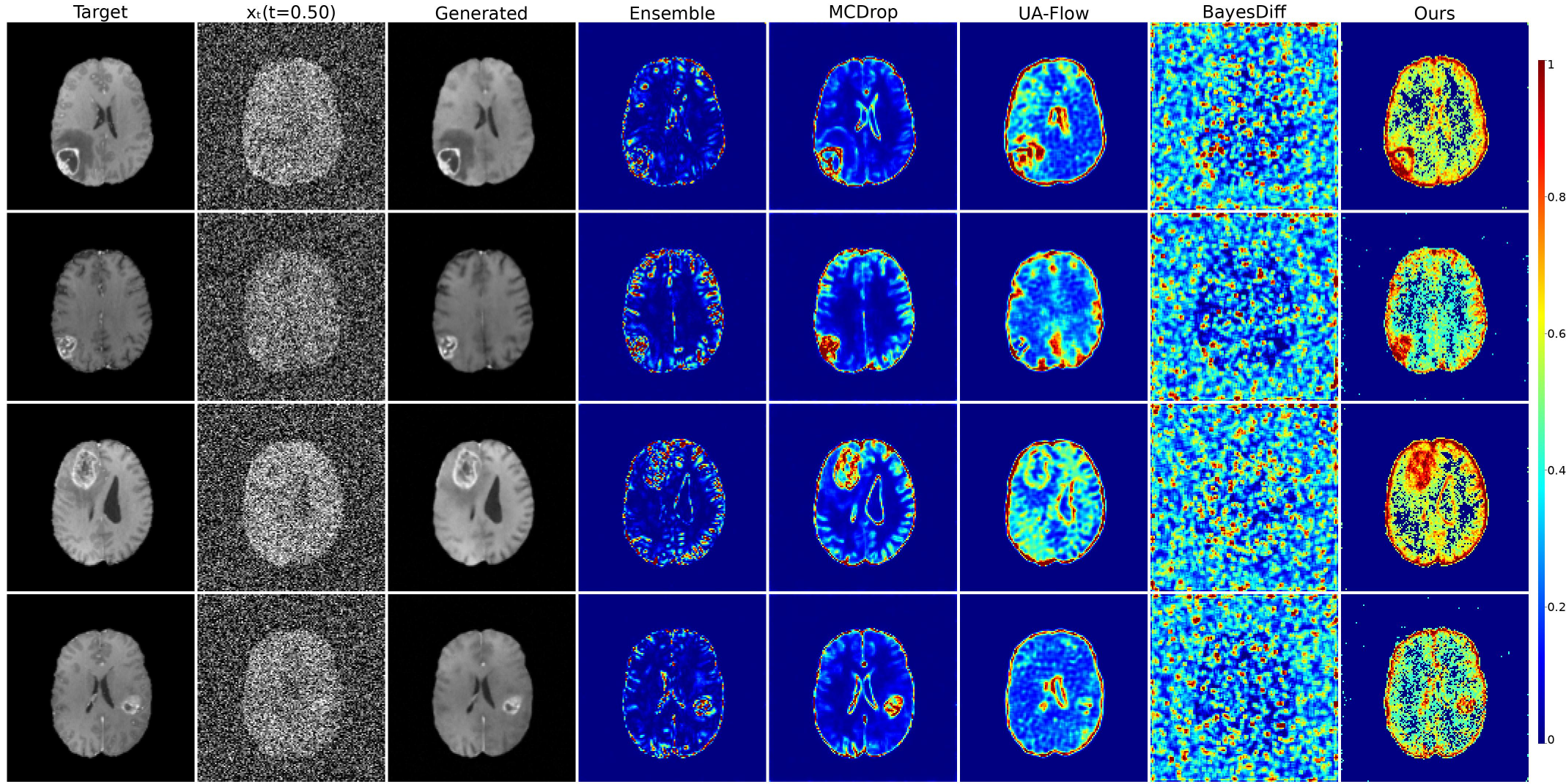}
    \caption{Per-pixel uncertainty maps on BraTS at interpolation
    time $t=0.5$.}
    \label{fig:brats-sup-t05}
\end{figure}

\begin{figure}[t]
    \centering
    \includegraphics[width=\linewidth]{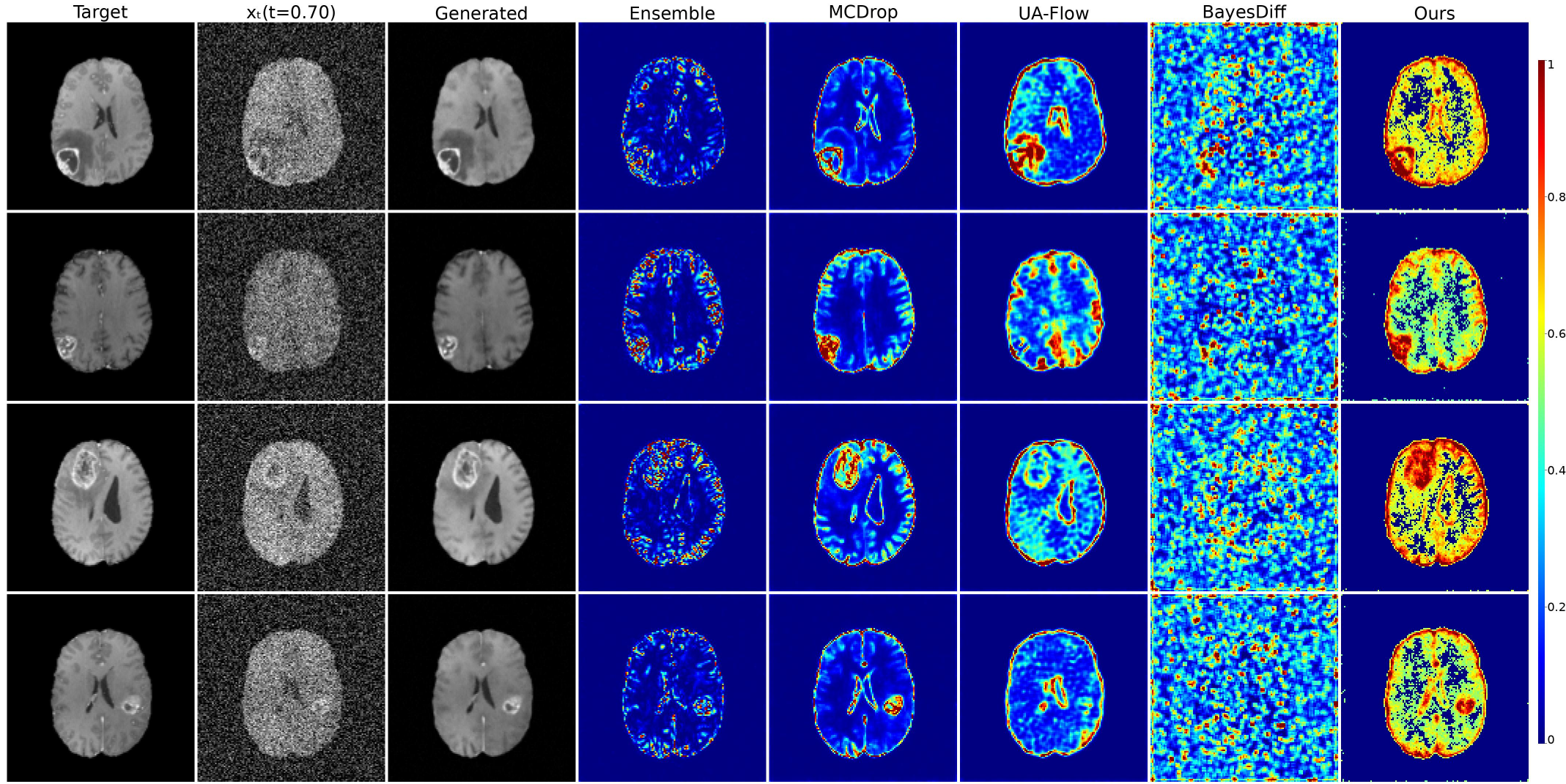}
    \caption{Per-pixel uncertainty maps on BraTS at interpolation
    time $t=0.7$.}
    \label{fig:brats-sup-t07}
\end{figure}

Figures~\ref{fig:cifar-sup-t03}, \ref{fig:cifar-sup-t05},
\ref{fig:cifar-sup-t07}, \ref{fig:brats-sup-t05}, and
\ref{fig:brats-sup-t07} present additional qualitative results
illustrating the variation in per-pixel uncertainty across
interpolation times on CIFAR-10 and BraTS. These examples provide
a more comprehensive view of the spatial uncertainty patterns for
different samples and demonstrate how the estimated uncertainty
evolves along the flow trajectory. 
All experiments were run on a single NVIDIA H200 NVL GPU.

\setcounter{topnumber}{5}
\setcounter{totalnumber}{5}
\renewcommand{\topfraction}{0.99}
\renewcommand{\textfraction}{0.01}
\renewcommand{\floatpagefraction}{0.50}
\makeatletter
\setlength{\@fptop}{0pt}
\setlength{\@fpsep}{14pt plus 0fil}
\setlength{\@fpbot}{0pt plus 1fil}
\makeatother

\end{document}